\documentclass{article}
\usepackage{booktabs}
\usepackage{multirow}
\usepackage{graphicx}
\usepackage{tabularx}
\usepackage{array}

\newcolumntype{Y}{>{\raggedright\arraybackslash}X}
\newcolumntype{C}{>{\centering\arraybackslash}X}

\usepackage[preprint]{neurips_2026}

\usepackage[utf8]{inputenc} 
\usepackage[T1]{fontenc}    
\usepackage{hyperref}       
\usepackage{url}            
\usepackage{booktabs}       
\usepackage{amsfonts}       
\usepackage{nicefrac}       
\usepackage{microtype}      
\usepackage{xcolor}         

\usepackage{amsmath}
\usepackage{amssymb}
\usepackage{mathtools}
\usepackage{amsthm}
\usepackage{mathrsfs}
\usepackage{bm}
\usepackage{multirow}
\usepackage[capitalize]{cleveref}

\newcommand{\Real}{\mathbb{R}}
\DeclarePairedDelimiter{\ceil}{\lceil}{\rceil}
\title{A Robust In-Context Model for Conservation Laws: Injecting Context into Flux Neural Operators via Recurrent Vision Transformers}

\author{%
  Taeyoung Kim\thanks{Equal contribution. Correspondence to: Taeyoung Kim <taeyoungkim@kias.re.kr>, Joon-Hyuk Ko <jhko725@kias.re.kr>.} \\
  Center for AI and Natural Sciences\\
  Korea Institute for Advanced Study\\
  Seoul, South Korea 02455 \\
  \texttt{taeyoungkim@kias.re.kr} \\
  \And
  Joon-Hyuk Ko\footnotemark[1]\\
  Center for AI and Natural Sciences\\
  Korea Institute for Advanced Study\\
  Seoul, South Korea 02455 \\
  \texttt{jhko725@kias.re.kr}
}

\begin{document}

\maketitle

\begin{abstract}
  We propose an architecture that augments the Flux Neural Operator (Flux NO), which combines the classical finite volume method (FVM) with neural operators, with ViT-based context injection. Our model is formulated as a hypernetwork: it extracts solution dynamics over a finite temporal window, encodes them with a recurrent Vision Transformer, and generates the parameters of a context-conditioned neural operator. This enables the model to infer and solve conservation laws without explicit access to the governing equation or PDE coefficients. Experimentally, we show that the proposed method preserves the robustness, generalization ability, and long-time prediction advantages of Flux NO over standard neural operators, while delivering reliable numerical solutions across a broad range of conservative systems, including previously unseen fluxes. Our code is available at \url{https://github.com/xx257xx/CONTEXT_FLUX_NO}.
\end{abstract}

\section{Introduction}

Neural-network-based methods for scientific computing have rapidly emerged as a major research direction, with a wide range of paradigms being introduced in quick succession. This evolution can be broadly understood as three successive shifts. First, physics-informed neural networks (PINNs) were proposed to solve partial differential equations (PDEs) by directly optimizing neural networks subject to the governing equations together with initial and boundary conditions~\citep{raissi2019}. Second, operator learning introduced a different perspective: rather than solving each PDE instance independently, neural operators learn the solution map of a prescribed PDE family, enabling direct prediction of forward or inverse solutions from input conditions~\citep{li2020,lu2021,kovachki2023}. More recently, inspired by the few-shot and in-context capabilities of Transformer-based foundation models~\citep{brown2020,dosovitskiy2020}, this viewpoint has been extended to scientific machine learning, giving rise to PDE foundation models that aim to solve diverse classes of PDEs by conditioning on contextual information such as observed dynamics, equation families, or domain structure~\citep{subramanian2023,hao2024,herde2024}.

Motivated by this line of development, we revisit the classical finite volume method (FVM), in which the evolution of conservation laws is governed by numerical fluxes at cell interfaces~\citep{leveque2002}, and combine it with neural operators in the spirit of the Flux Neural Operator (Flux NO)~\citep{kim2025a,kim2025b}. Building on this formulation, we propose a recurrent ViT-based context injection mechanism that lifts Flux NO into a in-context-model framework. The resulting model infers the underlying dynamics from short solution trajectories and adapts its numerical-flux operator accordingly, without requiring explicit knowledge of PDE coefficients or closed-form flux expressions.

Our main contributions are as follows:
\begin{itemize}
    \item We formulate an in-context flux-learning problem for parametric conservation laws, where a short observed trajectory is used to infer a latent numerical flux operator.
    \item We introduce a context-conditioned Flux Neural Operator in which a recurrent ViT encoder produces a compact context code that conditions the finite-volume flux operator.
    \item We show that enforcing a conservative flux-difference update improves autoregressive stability and OOD robustness compared with generic PDE foundation-model baselines on one-dimensional conservation-law benchmarks and a related diffusive Burgers-type problem.
\end{itemize}

\section{Background}
\label{sec:background}

This section reviews the ingredients that motivate our architecture. We emphasize two points. First, conservation laws require numerical updates that respect flux-difference structure, and Flux NOs encode this conservative structure, but are not inherently designed for in-context adaptation across unseen flux functions. Second, recent PDE foundation models provide context-conditioned adaptability, but often do so with generic prediction architectures that do not explicitly preserve conservative numerical structure.

\subsection{Conservation laws and Flux Neural Operators}

We consider conservation laws of the form
\begin{equation}
    \partial_t \bm{u}
    +
    \nabla \cdot \bm{F}(\bm{u};\bm{p})
    =
    0,
    \label{eq: conservation law}
\end{equation}
where \(\bm{u}(t,\bm{x})\in\mathbb{R}^{d}\) is the conserved state and \(\bm{F}(\bm{u};\bm{p})\) is the physical flux, possibly parameterized by coefficients \(\bm{p}\). The key structure of \cref{eq: conservation law} is that temporal evolution is completely determined by flux imbalance. 

In a one-dimensional finite volume discretization on \(N_x\) cells of width \(\Delta x\), the conservative update reads
\begin{equation}
    u^{\,n+1}_i=
    u^{\,n}_i-\frac{\Delta t}{\Delta x}
    \left(
        \hat{f}^{\,n}_{i+\frac12}
        -
        \hat{f}^{\,n}_{i-\frac12}
    \right); \quad i=0,\dots , N_x-1.
    \label{eq:fvm-update}
\end{equation}
where \(\hat{f}^{\,n}_{i+\frac12}\) is the numerical flux at interface \(i+\tfrac12\). This telescoping flux-difference structure ensures discrete conservation and is particularly important for nonlinear hyperbolic problems, where solutions develop shocks and long-time prediction requires stable transport behavior.

Many neural operator architectures \citep{li2020,lu2021} directly predict future solution fields and do note explicitly enforce \cref{eq:fvm-update}, which can lead to conservation errors or unstable error accumulation during autoregressive rollout. Flux Neural Operators \citep{kim2025a,kim2025b} address this by directly learning the numerical flux.

Given integers \(r\ge 0\) and \(s \ge 1\) defining the left and right stencil extents, the Flux NO predicts
\begin{equation}
    \hat{f}^{\,n}_{i+\frac12}=G_{\bm{\Theta}}\left(u_{i-r}^n,\dots,u_{i+s}^n\right),
    \label{eq:fluxno-interface-flux}
\end{equation}
where \(G_{\bm{\Theta}}\) is a neural operator acting on the \(r+s+1\)-cell stencil surrounding the interface \(i+\tfrac12\) and the next state is obtained by substituting \cref{eq:fluxno-interface-flux} into \cref{eq:fvm-update}. Thus, the model is constrained to evolve the solution through flux differences, giving it an inductive bias aligned with conservation laws. 

Different architectures for \(G_{\bm{\Theta}}\) have been used in the literature, such as MLPs or Fourier neural operators. In this work, we use an efficient 1D Convolution + MLP architecture instead, which we detail in \cref{subsec:fluxno-target}.

\subsection{Context conditioning for neural PDE solvers}

Recent work has begun to move from single-equation neural operators toward foundation models for PDEs. The goal is to train models that can operate across broader families of equations, coefficients, discretizations, and physical regimes by conditioning on contextual information, such as short observed trajectories, equation descriptors, simulation metadata, or prompt-like input--output examples. 

For the \emph{context-conditioning} task, a model observes a short trajectory of an unknown PDE instance must adapt its predictions accordingly, without explicit knowledge of the governing equation, PDE coefficients, or closed-form flux expressions. Several recent architectures pursue this goal through transformer-based sequence modeling, Fourier attention, or hypernetwork conditioning \citep{yang2024,yang2025,hao2024,morel2025}.

However, these methods employ generic prediction architectures - global operator regression or sequence-to-sequence models - that do not enforce the conservative flux-difference structure of \cref{eq:fvm-update}.
No existing method combines in-context adaptation with a conservative numerical backbone. The architecture we introduce next addresses this gap.

\section{In-Context Flux Neural Operator}
\label{section: model architecture}
Given a parametrized hyperbolic conservation law \cref{eq: conservation law}, our goal is to learn a context-conditioned
evolution operator that ingests a short context trajectory and predicts the state at the next timestep. 

Let \(\bm{u}^n\in\mathbb{R}^{d\times N_{\bm{x}}}\) denote the grid-sampled state at time \(t=n\Delta t\).
We seek to learn the mapping 
\[
    \bm{u}^{n-k+1:n}
    :=
    (\bm{u}^{n-k+1},\ldots,\bm{u}^{n})
    \in
    \mathbb{R}^{k\times d\times N_{\bm{x}}} \rightarrow \bm{u}^{n+1}\in\Real^{d\times N_x}.
\]

Rather than learning this map as an unconstrained input--output predictor, we decompose our model into three components: 
\begin{enumerate}
    \item \textbf{Context encoder} \(E\): compresses the input context trajectory into a context vector \(\bm{c} \in \Real^e\), implicitly solving an inverse problem to identify the parameters \(\bm{p}\) underlying the trajectory.
    \item \textbf{Hypernetwork} \(H\): maps \(\bm{c}\) to the weights of the target network \(\bm{\Theta}\in \Real^q\).
    \item \textbf{Target network} \(F\): a Flux NO with context dependent weights \(\bm{\Theta}\), that advances \(\bm{u}^n\) to \(\bm{u}^{n+1}\) via the conservative update of \cref{eq:fvm-update}.
\end{enumerate}

This imposes an information bottleneck
$\mathbb{R}^{k \times d \times N_x} \to \mathbb{R}^e \to \mathbb{R}^q$
with $e \ll q$.

The encoder does not receive the analytical flux function, PDE coefficients, or equation labels; all conditioning information must be inferred from the observed solution history.

\subsection{Context Encoder}
\label{subsec:encoder-hypernetwork}

The encoder must respect temporal causality as the progression of the solution over time carries essential information about wave speeds, dissipation, and nonlinear interactions that identify the underlying PDE. We therefore adopt a temporally recurrent Vision Transformer design inspired by TRecViT~\citep{patraucean2025}, where temporal mixing is handled by gated linear recurrent units~\citep{de2024,botev2024} and spatial mixing by vision transformer blocks.

\paragraph{Temporal recurrent mixing and spatial attention.}
The encoder is designed to process temporal and spatial axes separately while respecting causality along time. 

Given \(\bm{u}^{n-k+1:n}\in\mathbb{R}^{k\times d\times N_{\bm{x}}}\), the encoder first tokenizes each time slice using a ViT patch embedding with learnable positional encodings~\citep{dosovitskiy2020}:
\begin{equation}
    \bm{v}^{(0)}
    =
    \mathrm{PatchEmbed}(\bm{u}^{n-k+1:n})
    \in
    \mathbb{R}^{k\times P\times e},
\end{equation}
where \(P\) is the number of spatial patches. Each encoder layer alternates between temporal recurrent mixing for each spatial token and spatial self-attention for each time step:
\begin{equation}
    \widehat{\bm{v}}^{(\ell)}_{:,p}
    =
    \mathrm{TemporalBlock}^{(\ell)}
    \left(
        \bm{v}^{(\ell)}_{:,p}
    \right),
    \qquad
    \bm{v}^{(\ell+1)}_{t,:}
    =
    \mathrm{SpatialTransformer}^{(\ell)}
    \left(
        \widehat{\bm{v}}^{(\ell)}_{t,:}
    \right).
\end{equation}
In our implementation, the temporal block is a residual recurrent block based on a gated linear recurrent unit with a causal depthwise one-dimensional convolution. This alternating structure allows the encoder to propagate information through the observed trajectory while modeling spatial interactions at each time step.

After the final layer, we apply token-wise layer normalization and average the final temporal state over spatial tokens:
\begin{equation}
    \bm{c}
    =
    \frac{1}{P}
    \sum_{p=1}^{P}
    \mathrm{LayerNorm}
    \left(
        \bm{v}^{(L)}_{k,p}
    \right)
    \in
    \mathbb{R}^{e}.
\end{equation}

\subsection{Hypernetwork}
The hypernetwork is a two-layer MLP that maps the context vector to the full parameter vector of the target network:
\begin{equation}
    H(\bm{c})=W_{out}\cdot\mathrm{GeLU}(W_{in}\bm{c}+\bm{b}_{in})+\bm{b}_{out},
\end{equation}
where \(W_{in}\in\Real^{h\times e}, \bm{b}_{in}\in\Real^{e}, W_{out}\in\Real^{q\times h}, \bm{b}_{out}\in\Real^{q}\).

Since $q \sim \mathcal{O}(10^6)$, storing $W_{\mathrm{out}}$ as a dense matrix is prohibitive; we instead use a block-diagonal structure with $B$ blocks, reducing its parameter count from $qh$ to $qh/B$.

For initialization, we use Bias-Hyperinit \citep{beck2023}: \(W_{\mathrm{out}}\) is initialized to zero and \(\bm{b}_{\mathrm{out}}\) is set to the concatenation of the target network's default-initialized parameters, so that \(H(\bm{c}) = \bm{b}_{\mathrm{out}}\) at initialization
regardless of the input \(\bm{c}\).

\subsection{Flux Neural Operator Target Network}
\label{subsec:fluxno-target}

The target network is a Flux~NO whose parameters \(\bm{\Theta}\) are generated by the hypernetwork. Each forward pass predicts numerical fluxes at all cell interfaces and advances the state by one conservative time step. We first describe the one-dimensional case and then extend to higher dimensions.

\paragraph{Stencil construction.}
As shown in \cref{eq:fluxno-interface-flux}, the numerical flux at interface \(i+\tfrac12\) depends on the \(r+s+1\)-cell stencil \((u_{i-r}^n,\dots , u_{i+s}^n)\). Leveraging that this domain of dependence corresponds to that of a 1D convolution, we compute the following:
\begin{equation}
    \bm{z}^{(0)} = \mathrm{Conv1D}\left(
    \mathrm{Pad}(\tilde{\bm{u}}^n;\; \text{left}=r+1, \text{right}=s);\; \text{kernel}=w
  \right)
  \;\in\; \mathbb{R}^{d_{\mathrm{lift}} \times (N_x+1)},
  \label{eqn: flux no conv1d}
\end{equation}
where \(\tilde{\bm{u}}^n\in\Real^{(d+d_g)\times N_x}\) is the input \(\bm{u}^n\) with the grid coordinates appended an additional channel, and padding is performed consistent with the boundary conditions \footnote{For periodic boundaries, wrap-around (circular) padding is used.}. Here, $d_g$ denotes the number of grid-coordinate channels appended to the input \footnote{E.g.\ $d_g=1$ for one spatial dimension.}.

The resulting \(\bm{z}^{(0)}\) has length \(N_x+1\) in its spatial dimension, corresponding to the cell interfaces \((-\tfrac12,\dots N_x-\tfrac12)\).

\paragraph{Pointwise MLP and conservative update.} To enhance the expressivity of the target function, we apply a shared \(L\)-layer MLP with GeLU nonlinearities and output dimension of \(d\) on each spatial entry of \(\bm{z}^{(0)}\):
\begin{equation}
    \bm{z}^{(L)}_i = MLP(\bm{z}^{(0)}_i)\in \Real^d, \quad i=0,..,N_x.
    \label{eqn: flux no mlp}
\end{equation}
Here, \(\bm{z}^{(L)}_i\) represents the model prediction for the numerical flux at interface \(i-\tfrac12\). This allows the solution to be advanced in time as 
\begin{equation}
    \bm{u}^{n+1}=\bm{u}^{n}-\frac{\Delta t}{\Delta x}
    \left(
        \bm{z}^{(L)}_{1:N_x}-\bm{z}^{(L)}_{0:N_x-1}
    \right),
    \label{eq:context-fluxno-vector-update}
\end{equation}
corresponding to a vectorized version of \cref{eq:fvm-update}.

\paragraph{Extension to higher dimensions.} For hyperbolic conservation laws with additional spatial dimensions, we employ a dimensional-splitting approach. For 2D, this corresponds to two separate 1D Flux NOs \(G^x_{\bm{\Theta}_x}\) and \(G^y_{\bm{\Theta}_y}\) - each operates on stencils along its own axes, and produces \(x\) or \(y\) direction numerical flux according to \cref{eqn: flux no conv1d,eqn: flux no mlp}.

The conservative update becomes
\begin{equation}
    u_{i,j}^{n+1}
  = u_{i,j}^n
    - \frac{\Delta t}{\Delta x}\left(\hat{f}^x_{i+\frac{1}{2},j} - \hat{f}^x_{i-\frac{1}{2},j}\right)
    - \frac{\Delta t}{\Delta y}\left(\hat{f}^y_{i,j+\frac{1}{2}} - \hat{f}^y_{i,j-\frac{1}{2}}\right).
\end{equation}
Both Flux NOs share the same hypernetwork, and hence the same context vector; their parameters correspond to disjoint slices of \(\bm{\Theta}\).

\begin{figure*}[h]
  \centering
  \includegraphics[width=1.0\textwidth]{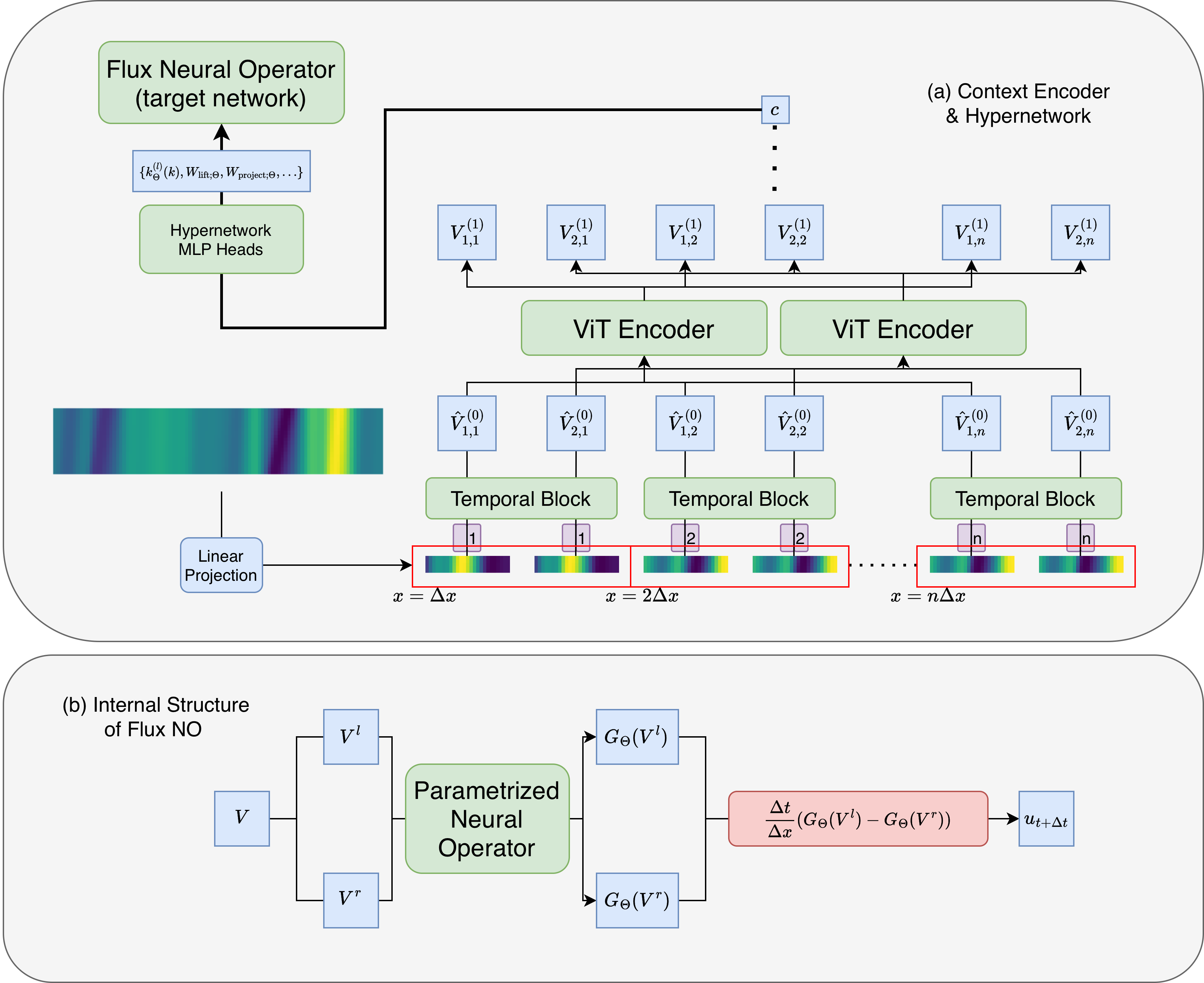}
  \caption{
  Overview of HFluxNO. 
  (a) A temporally recurrent Vision Transformer encodes the context trajectory by alternating temporal recurrent mixing and spatial attention, producing a context vector that is mapped by a hypernetwork to Flux NO parameters.
  (b) The generated Flux NO target network predicts numerical fluxes, which are used in a conservative finite-volume update.
  }
  \label{fig: vit backbone}
\end{figure*}

We illustrate the overall architecture in Figure~\ref{fig: vit backbone}. We also refer the readers to \cref{appendix: further architecture details} for further details and rationale on the architecture design.

\section{Experiments}
\subsection{Baselines}
To demonstrate the efficacy of our method, we selected the following state-of-the-art models from recent literature.  For a fair performance comparison, we implement all models in \texttt{JAX} \cite{jax2018} porting over original implementations if necessary. We provide a brief description of the baselines below, and refer readers to \cref{modelconfigs} for additional details. Unlike the other baselines, ICON is fundamentally a snapshot-based
model: given a set of in-context example pairs, it predicts the target
solution in a single shot rather than rolling the state forward
autoregressively in time. Because of this, ICON cannot be evaluated on a
footing directly comparable to the autoregressive baselines considered
here, and reporting it alongside them in the main text would not amount to
a fair comparison. A dedicated
comparison of the snapshot and autoregressive paradigms, together with the
corresponding results on the 1D benchmarks, is provided in
\cref{appendix:baselines}.

\paragraph{ICON} \citep{yang2024,yang2025} is a decoder-only transformer language model that was repurposed for operator learning. Instead of language tokens, the model is trained to ingest the PDE solution field sampled at discrete time points and generate the output at some future point in time.

\paragraph{DPOT} \citep{hao2024} is a non-transformer model that first compresses the input trajectory using spatial patch embedding, followed by a learnable weighted sum along the time axis. Subsequently, Fourier attention layers \citep{guibas2021,hao2024} are applied on the aggregated context to learn kernel integral transforms conditional to the input context.

\paragraph{DISCO} \citep{morel2025} is another hypernetwork-based architecture, with the axial vision transformer architecture from \citet{mccabe2024} as the hypernetwork, and a neural ordinary differential equation \citep{chen2018,kidger2022} with a U-Net \citep{ronneberger2015} vector field as the target network. Next time-step predictions are generated by numerically integrating the U-Net vector field using an adaptive Runge-Kutta solver, which can make this method more computationally expensive than its counterparts.

\subsection{Datasets}
While large, high quality PDE datasets have been made available in recent years \citep{takamoto2022,ohana2024,koehler2024}, many of these datasets have limited variety in the equational form of the PDEs, their coefficient values, and the function family the initial conditions are sampled from. While a natural consequence of the difficulty of generating high quality PDE solutions, this limitation makes it difficult to gauge the generalization capabilities of the trained multiphysics neural operators. As such, we perform experiments with newly generated datasets designed to test both
in-distribution performance and controlled forms of out-of-distribution generalization. For newly generated data, we provide a brief description of each dataset below, and provide in-depth simulation details in \cref{appendix: data generation}.

\paragraph{1D Cubic Conservation Laws}
We first consider the problem of learning a family of 1D cubic conservation laws, as proposed by \citet{yang2024}. The governing equation is given as,
\[
u_t + (c_1 u + c_2 u^2 + c_3 u^3)_x = 0,\quad x\in [0,1],\quad (c_1,c_2,c_3)\sim \mathrm{Unif}([-1,1]^3)
\]
with periodic boundary conditions and sample the initial conditions \(u(0)\) from a 1D Gaussian random field with the periodic covariance function.  


\paragraph{1D Shallow Water Equations}
Next, we consider a parametrized form of the 1D shallow-water equations. 
Let \(m=hu\) denote the momentum. The state is \(q=(h,m)^\top\), and the governing equation is
\begin{equation}
    q_t + F(q)_x = 0,\qquad x\in[0,1],
\end{equation}
with flux
\begin{equation}
    F(q)
    =
    \begin{pmatrix}
        \alpha m \\
        \gamma m^2/h + \frac{1}{2}\beta h^2
    \end{pmatrix},
    \qquad
    (\alpha,\gamma,\beta)
    \sim
    \mathrm{Unif}
    \left(
        [0.5,1.5]\times[0.5,1.5]\times[8,12]
    \right).
\end{equation}
The standard shallow-water equations correspond to \((\alpha,\gamma,\beta)=(1,1,g)\).
For the initial conditions, we sample \(m(0)\) from a Gaussian random field and \(h(0)\)
from a lognormal random field to ensure that the water height remains non-negative.

\paragraph{1D Viscous Burgers Equation} The last equation we simulate is the viscous Burgers equation, given as
\begin{equation}
    u_t+(a\cdot u^2)_x=\nu u_{xx}, \quad x\in[0,1],\quad (a,\nu)\sim\text{Unif}([0.5,1.5]\times[0.005,0.015]).
\end{equation}
Note that this equation is not a conservation law (\cref{eq: conservation law}) due to the presence of a dissipative term on the right hand side. Therefore we include this dataset in our benchmarks to gauge if our HFluxNO model can handle more general cases beyond the strictly conservative setting it was motivated by.

For all simulated datasets, the equations are solved in the time interval \(t\in[0,0.4]\) with a sampling period of \(\Delta t=0.005\). We generate 100 initial conditions per coefficient choice, and 1000, 100, 100 coefficient choices for the training, validation, and test datasets respectively.

\subsection{Model training and evaluation} 
All models were trained using the mean squared error between model predictions and data for a single time step prediction. We set the context length to be \(k=20\) for most of our experiments. All models were trained for 50000 gradient steps, with the AdamW optimizer and a linear warm-up cosine decay schedule for the learning rate.

We evaluate the trained models on (i) in-distribution accuracy, (ii) out-of-distribution robustness (shock-dominated regimes, sine fluxes),
(iii) long-time roll-out beyond the training horizon.
For evaluation metric, we used the relative $l^{2}$ and $l^{\infty}$ norms which are defined as in \eqref{relloss}
\begin{equation}
        \label{relloss}
        \text{Rel. } l^{2}(u,u_\text{target}):=\frac{\|u-u_{\text{target}}\|_{2}}{\|u_{\text{target}}\|_{2}}, \quad
        \text{Rel. } l^{\infty}(u,u_\text{target}):=\frac{\|u-u_{\text{target}}\|_{\infty}}{\|u_{\text{target}}\|_{\infty}} 
\end{equation}
over space at each time, and then averaged over time; or over the full spatiotemporal grid.

\section{Results}

\subsection{In-distribution predictions.} We first evaluate the trained models under the in-distribution setting, using a test dataset of 100 coefficient combinations with 100 initial conditions each, sampled from the same distributions as the training data. We consider two types of model predictions - (i) a single step forecast, where the context is fed into the model to predict the solution \(\Delta t\) time later, and (ii) a short autoregressive rollout, where the model output is recursively fed back into the model as input context for 20 times to generate a short prediction trajectory over a time horizon of \(20\Delta t\).


\begin{table*}[htp!]
\centering
\caption{In-distribution (ID) prediction accuracy on the 1D benchmark datasets.
Reported as mean \(\pm\) std over three training runs. The best results are in bold, and the runner-ups are underlined.}
\label{tab: cubid_1d_in_distribution}
\setlength{\tabcolsep}{7pt}
\renewcommand{\arraystretch}{1.15}
\resizebox{\textwidth}{!}{\begin{tabular}{llcccc}
\toprule
& & \multicolumn{2}{c}{Single step} & \multicolumn{2}{c}{Autoregressive Rollout (20 steps)} \\
\cmidrule(lr){3-4}\cmidrule(lr){5-6}
Dataset & Model & Rel.~$l^2\ (\downarrow)$  & Rel.~$l^\infty\ (\downarrow)$ & Rel.~$l^2\ (\downarrow)$  & Rel.~$l^\infty\ (\downarrow)$\\
\midrule
\multirow{3}{*}{Cubic} & DPOT & \underline{7.40e-3$\pm$9.09e-5} & \underline{2.42e-2$\pm$3.84e-4} & 1.36e-1\(\pm\)3.12e-3 & 6.78e-1\(\pm\)9.65e-3\\
& DISCO & 1.06e-2$\pm$1.99e-3 & 3.90e-2$\pm$9.65e-3 & \underline{8.22e-2\(\pm\)5.58e-3} & \underline{4.55e-1\(\pm\)3.12e-2} \\

\cmidrule(lr){2-6}
& HFluxNO  & \textbf{4.10e-3$\pm$1.15e-4} & \textbf{1.55e-2$\pm$6.17e-4} & \textbf{5.21e-2\(\pm\)2.15e-3} & \textbf{3.68e-1\(\pm\)1.56e-2} \\
\midrule
\multirow{3}{*}{Shallow water} & DPOT & \underline{8.49e-3$\pm$4.00e-4} & \underline{2.88e-2$\pm$1.59e-3} & \underline{1.05e-1\(\pm\)3.66e-3} & 4.63e-1\(\pm\)1.00e-2\\
& DISCO & 1.42e-2$\pm$6.28e-4 & 5.07e-2$\pm$1.94e-3 & 1.11e-1\(\pm\)5.82e-3 & \underline{4.45e-1\(\pm\)2.01e-2} \\

\cmidrule(lr){2-6}
& HFluxNO  & \textbf{6.27e-3$\pm$2.71e-5} & \textbf{2.29e-2$\pm$3.81e-4} & \textbf{7.55e-2\(\pm\)1.26e-3} & \textbf{3.98e-1\(\pm\)4.21e-3} \\
\midrule

\multirow{3}{*}{Viscous Burgers} & DPOT & \underline{2.63e-3$\pm$2.93e-6} & \underline{7.17e-3$\pm$4.29e-5} & \underline{4.66e-2\(\pm\)1.07e-3} & 1.56e-1\(\pm\)4.39e-3\\
& DISCO & 3.43e-3$\pm$2.09e-4 & 9.65e-3$\pm$6.01e-4 & 6.56e-2\(\pm\)3.35e-2 & \underline{3.43e-1\(\pm\)2.34e-1} \\

\cmidrule(lr){2-6}
& HFluxNO  & \textbf{1.32e-3$\pm$2.90e-5} & \textbf{3.44e-3$\pm$7.30e-5} & \textbf{1.72e-2\(\pm\)2.35e-4} & \textbf{6.46e-2\(\pm\)6.51e-4} \\
\bottomrule
\end{tabular}}
\end{table*}

From the results in \cref{tab: cubid_1d_in_distribution}, we find that the models often have markedly higher relative \(l^\infty\) errors than relative \(l^2\) errors, which stems from the difficulty of exactly capturing shock front locations over time, as opposed to getting the overall form of the solution correctly. The baseline models show an interesting trend: DISCO performs worse than DPOT in several single-step settings, but its stronger dynamical prior improves longer autoregressive rollouts. However, incorporating prior structure into the model pays off in the longer term, with DISCO outperforming the other more flexible baselines.

In contrast to this trade-off between single-step and autoregressive performance for the baseline models, we find that HFluxNO consistently outperforms the baselines in both single-step prediction and autoregressive rollout. This indicates that the choice of the model prior structure also greatly matters, and that our architecture design based on the finite volume method is highly effective in learning hyperbolic conservation laws.

\begin{figure*}[htp!]
  \centering
  \includegraphics[width=\textwidth]{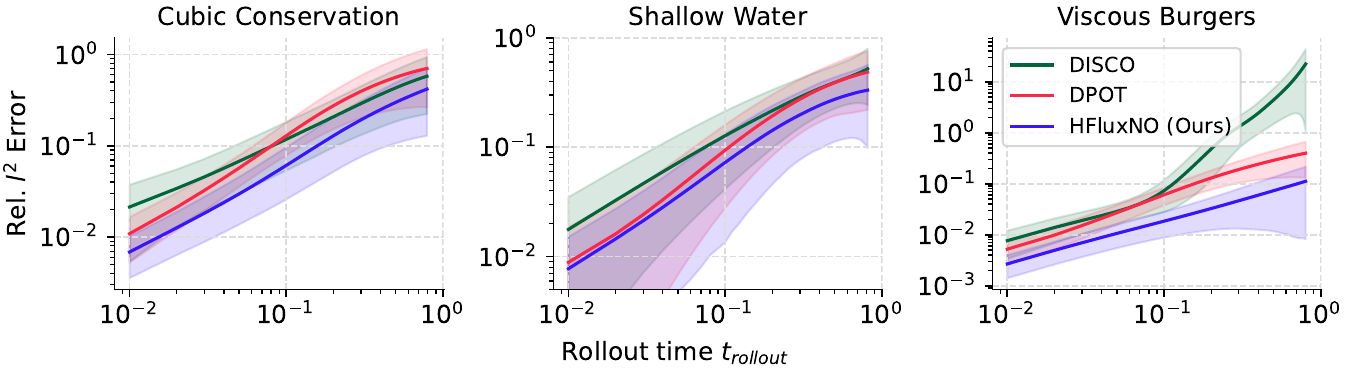}
  \caption{
  Long time prediction performances of the in-context neural operator models across different datasets.
  }
  \label{fig: long time predictions}
\end{figure*}

\paragraph{Long time prediction capabilities}
To further stress test the predictive capabilities of the trained models, we generated long time predictions corresponding to a rollout time of \(t_{rollout}=0.4\). From the results shown in \cref{fig: long time predictions} we see that our model consistently maintains lower error over time compared with the baselines. Furthermore, we see that the way error accumulates in the model predictions over time differs (\cref{fig: long time predictions}, right panel). DPOT and, to a lesser extent, DISCO quickly accumulate high-frequency artifacts with increasing rollouts, which is a well-known problem plaguing autoregressive neural operator architectures \citep{lippe2023,worrall2024}. In contrast, our model does not suffer from such artifacts, with errors only stemming from a slight misprediction of the wave propagation speed. This indicates that our model has properly learned the local physics of the problem, due to the effectiveness of the built-in inductive biases. 

\paragraph{Out-of-distribution generalization.}

We next evaluate the out-of-distribution (OOD) generalization capability of the trained models by (i) assessing model performances on a new dataset, generated using a different, shock-dominated initial condition distribution. Additionally, for the cubic conservation law experiment, We further test models on (ii) seen initial conditions (GRFs) but unseen equations (sine-flux dynamics), and (iii) unseen initial conditions and equations. 

\begin{figure*}[htp!]
  \centering
  \includegraphics[width=\textwidth]{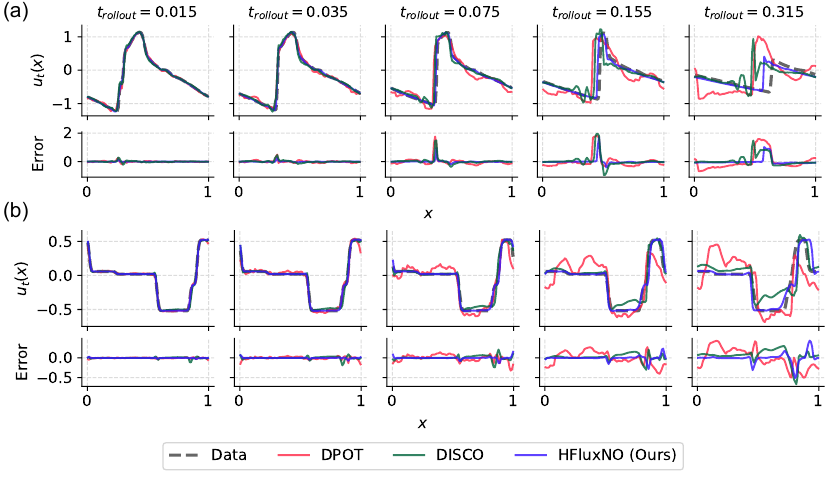}
  \caption{
Qualitative rollout examples. The top row (a) shows an in-distribution cubic test trajectory,
whereas the bottom row (b) shows an OOD trajectory with shock-dominated initial conditions
and sine-flux dynamics.
}
  \label{fig: trajectory rollouts}
\end{figure*}

These settings respectively test robustness to shifted initial-condition distributions, and generalization to a different flux family. Datasets were generated analogously to the in-distribution dataset (details are provided in \cref{appendix: data generation}) and models were evaluated directly on these OOD test sets without fine-tuning. The quantitative results for these OOD settings are reported in Table~\ref{tab: ood_results}, and qualitative examples are shown in Figure~\ref{fig: trajectory rollouts}.

\begin{table*}[htp!]
\centering
\caption{OOD generalization performance. The Cubic, Shallow Water, and Viscous Burgers rows use shock-dominated initial conditions. The sine-flux rows evaluate cubic-trained models on unseen sine-flux dynamics without retraining, either with GRF initial conditions or shock-dominated initial conditions. Reported as mean \(\pm\) std over three training runs.}
\label{tab: ood_results}
\setlength{\tabcolsep}{7pt}
\renewcommand{\arraystretch}{1.15}
\resizebox{\textwidth}{!}{\begin{tabular}{llcccc}
\toprule
& & \multicolumn{2}{c}{Single step} & \multicolumn{2}{c}{Autoregressive Rollout (20 steps)} \\
\cmidrule(lr){3-4}\cmidrule(lr){5-6}
Dataset & Model & Rel.~$l^2\ (\downarrow)$  & Rel.~$l^\infty\ (\downarrow)$ & Rel.~$l^2\ (\downarrow)$  & Rel.~$l^\infty\ (\downarrow)$\\
\midrule
\multirow{3}{*}{Cubic} & DPOT & \underline{7.24e-3$\pm$1.99e-4} & \underline{2.45e-2$\pm$1.72e-4} & 1.58e-1\(\pm\)3.95e-3 & 8.23e-1\(\pm\)3.82e-2\\
& DISCO & 9.70e-3$\pm$1.07e-3 & 3.73e-2$\pm$4.19e-3 & \underline{8.58e-2\(\pm\)5.69e-3} & \underline{4.73e-1\(\pm\)2.63e-2} \\

\cmidrule(lr){2-6}
& HFluxNO  & \textbf{5.62e-3$\pm$2.74e-4} & \textbf{2.27e-2$\pm$1.45e-3} & \textbf{6.68e-2\(\pm\)1.69e-3} & \textbf{4.35e-1\(\pm\)8.96e-3} \\

\midrule
\multirow{3}{*}{Sine (with GRF)} & DPOT & 5.98e-3$\pm$9.64e-5 & \underline{1.16e-2$\pm$2.68e-4} & 1.64e-1\(\pm\)7.44e-3 & 4.92e-1\(\pm\)1.27e-2\\
& DISCO & \underline{5.90e-3$\pm$1.01e-3} & 1.25e-2$\pm$2.07e-3 & \underline{6.67e-2\(\pm\)6.96e-3} & \underline{2.33e-1\(\pm\)2.47e-2} \\

\cmidrule(lr){2-6}
& HFluxNO  & \textbf{2.66e-3$\pm$3.17e-5} & \textbf{6.44e-3$\pm$3.74e-5} & \textbf{4.14e-2\(\pm\)1.17e-3} & \textbf{1.78e-1\(\pm\)4.24e-3} \\
\midrule
\multirow{3}{*}{Sine } & DPOT & \underline{7.83e-3$\pm$1.99e-4} & \underline{2.74e-2$\pm$9.22e-4} & 1.76e-1\(\pm\)4.69e-3 & 9.29e-1\(\pm\)4.45e-2\\
& DISCO & 9.51e-3$\pm$1.99e-3 & 3.88e-2$\pm$7.78e-3 & \underline{9.01e-2\(\pm\)8.08e-3} & \underline{5.39e-1\(\pm\)2.21e-2} \\

\cmidrule(lr){2-6}
& HFluxNO  & \textbf{4.85e-3$\pm$2.85e-4} & \textbf{1.98e-2$\pm$1.53e-3} & \textbf{6.06e-2\(\pm\)2.02e-3} & \textbf{3.83e-1\(\pm\)7.62e-3} \\
\midrule
\multirow{3}{*}{Shallow water} & DPOT & \underline{7.45e-2$\pm$5.36e-3} & \underline{3.46e-1$\pm$2.78e-2} & 4.59e-1\(\pm\)6.07e-3 & 9.77e-1\(\pm\)5.47e-3\\
& DISCO & \textbf{5.85e-2$\pm$1.16e-3} & \textbf{2.46e-1$\pm$4.29e-3} & \textbf{3.35e-1\(\pm\)1.01e-2} & \underline{9.40e-1\(\pm\)6.75e-2} \\

\cmidrule(lr){2-6}
& HFluxNO  & 1.25e-1$\pm$2.83e-3 & 4.96e-1$\pm$6.25e-3 & \underline{4.28e-1\(\pm\)6.62e-3} & \textbf{8.47e-1\(\pm\)1.28e-2} \\
\midrule

\multirow{3}{*}{Viscous Burgers} & DPOT & \underline{1.89e-3$\pm$2.10e-5} & \textbf{5.08e-3$\pm$1.67e-4} & \underline{4.50e-2\(\pm\)3.19e-3} & \underline{1.78e-1\(\pm\)1.17e-2}\\
& DISCO & 2.73e-3$\pm$9.74e-5 & 7.58e-3$\pm$4.24e-4 & 4.53e-2\(\pm\)1.41e-2 & 2.41e-1\(\pm\)1.25e-1 \\

\cmidrule(lr){2-6}
& HFluxNO  & \textbf{1.64e-3$\pm$2.43e-5} & \underline{5.24e-3$\pm$7.20e-5} & \textbf{2.21e-2\(\pm\)5.04e-4} & \textbf{9.99e-2\(\pm\)9.47e-4} \\
\bottomrule
\end{tabular}}
\end{table*}

\paragraph{Limitations.}
Our experiments focus primarily on one-dimensional conservation-law dynamics, with one additional diffusive Burgers-type benchmark beyond the strictly conservative setting. Although the proposed architecture is motivated by a conservative finite-volume structure, its performance on higher-dimensional systems, complex geometries, strongly coupled multiphysics problems, and real-world noisy observations remains to be investigated. In addition, the current study evaluates context adaptation on selected equation families, and broader generalization across substantially different PDE classes is left for future work.
 
\section{Conclusion}

In this work, we proposed HFluxNO, which extends Flux NO into a context-adaptive model for conservation-law dynamics. To handle temporal causality in the input trajectory, we designed a context-injection encoder hypernetwork based on a temporally recurrent Vision Transformer, while using the original Flux NO architecture as the target network. This design allows the model to infer latent governing dynamics from short solution histories and instantiate a context-conditioned conservative flux operator.

Through training and evaluation against recent baseline models, HFluxNO showed competitive or improved performance across several settings, including in-distribution prediction, out-of-distribution generalization with respect to initial conditions and flux functions, and long-time autoregressive prediction. The benchmark problems considered in this paper include one-dimensional scalar conservation laws, one-dimensional vector-valued conservation laws, and a viscous Burgers-type equation with an explicit diffusive term. These results suggest that combining in-context adaptation with a conservative flux-difference inductive bias can be beneficial for neural solvers of conservation-law dynamics.

Future work will extend this framework to richer multiphysics settings, including higher-dimensional systems, more diverse equation families, and more complex physical regimes.

\bibliography{references}
\bibliographystyle{abbrvnat}

\newpage
\appendix

\section{Additional baselines}
\label{appendix:baselines}

\subsection{ICON}

For ICON, we conducted experiments using the code from \cite{yang2024}, adopting the original experimental setup and protocol as closely as possible. Since the baseline models are designed to predict the solution after $\Delta t=0.005$, we trained one ICON model to predict the same forward time interval. In addition, following the original setting of \cite{yang2024}, we trained another ICON model to predict the solution after $\Delta t=0.1$. We denote these two models as ICON ($\tau=0.005$) and ICON ($\tau=0.1$), respectively.

The performance on the in-distribution test dataset is reported in \cref{tab:icon_id_cubic_tau_0005}. For ICON ($\tau=0.005$), we performed autoregressive rollout by fixing the randomly sampled context and recursively feeding the model output back as input. As shown in the table, this leads to very poor rollout performance. In contrast, ICON ($\tau=0.1$) predicts the target state with a single inference step and performs better than the former setting, but it still substantially lags behind the baseline models for repeated inferences. Since ICON ($\tau=0.005$) is effectively not meaningful for long-horizon prediction, we conducted the long-time prediction and OOD test experiments using ICON ($\tau=0.1$). The corresponding results are summarized in Table~\ref{tab:icon_generalization}.

The results for the viscous Burgers problem and the shallow water context problem are reported in Table~\ref{tab:icon_otherproblems}. For each problem, we evaluate ICON on both the in-distribution test set and the OOD test set with shock-dominated initial conditions. We remark that, in the viscous Burgers experiments, ICON occasionally produced divergent solution values during a single inference step. These divergent cases were excluded from the computation of the reported errors, suggesting a potential inference-time instability of ICON in this setting.

\begin{table*}[htp!]
\centering
\small
\caption{In-distribution prediction accuracy of ICON on the 1D cubic conservation law with $\tau=0.005$ and $\tau=0.1$ models.
We report single-step prediction errors and autoregressive rollout errors after $\Delta t=0.1$ over 20 steps for $\tau=0.005$, and two-step prediction $\Delta t=0.2$ for $\tau=0.1$.
Reported as mean \(\pm\) std over three training runs.}
\vspace{0.2cm}
\label{tab:icon_id_cubic_tau_0005}
\setlength{\tabcolsep}{7pt}
\renewcommand{\arraystretch}{1.15}

\begin{tabular}{lcccc}
\toprule
& \multicolumn{2}{c}{Single step}
& \multicolumn{2}{c}{Autoregressive Rollout (20 steps / 2 steps)} \\
\cmidrule(lr){2-3}\cmidrule(lr){4-5}
Model
& Rel.~$l^2$ ($\downarrow$)
& Rel.~$l^\infty$ ($\downarrow$)
& Rel.~$l^2$ ($\downarrow$)
& Rel.~$l^\infty$ ($\downarrow$)\\
\midrule
ICON ($\tau=0.005$)
& 1.38e-2 $\pm$ 2.52e-3
& 3.73e-1 $\pm$ 5.34e-2
& 1.06 $\pm$ 1.04e-1
& 1.33 $\pm$ 7.17e-2\\
\midrule
ICON ($\tau=0.1$)
& 1.05e-2 $\pm$ 1.95e-1
& 3.54e-1$\pm$ 5.81e-2
& 2.43e-1$ \pm$ 9.38e-2
& 8.06 e-2 $\pm$ 5.81e-2 \\
\bottomrule
\end{tabular}
\end{table*}

\begin{table*}[htp!]
\centering
\small
\caption{Generalization performance of ICON ($\tau=0.1$) on the 1D conservation laws.
We report OOD generalization under shock-dominated initial conditions, sine-flux dynamics, and their combination.
Reported as mean \(\pm\) std over three training runs.}
\vspace{0.2cm}
\label{tab:icon_generalization}
\setlength{\tabcolsep}{6pt}
\renewcommand{\arraystretch}{1.15}
\resizebox{\textwidth}{!}{
\begin{tabular}{lcccccc}
\toprule
& \multicolumn{2}{c}{Shock} 
& \multicolumn{2}{c}{Sine flux} 
& \multicolumn{2}{c}{Shock + sine flux} \\
\cmidrule(lr){2-3}
\cmidrule(lr){4-5}
\cmidrule(lr){6-7}
Model 
& Rel.~$l^2$ ($\downarrow$)
& Rel.~$l^\infty$ ($\downarrow$)
& Rel.~$l^2$ ($\downarrow$)
& Rel.~$l^\infty$ ($\downarrow$)
& Rel.~$l^2$ ($\downarrow$)
& Rel.~$l^\infty$ ($\downarrow$)\\
\midrule
ICON ($\tau=0.1$)
& 6.47e-2$\pm$3.50e-2 
& 6.33e-1$\pm$5.14e-1
& 6.29e-3$\pm$1.80e-3
& 9.35e-2$\pm$5.39e-2
& 4.37e-2$\pm$3.55e-2
& 6.43e-1$\pm$1.69e-1 \\
\bottomrule
\end{tabular}
}
\end{table*}

\begin{table*}[htp!]
\centering
\caption{In-distribution and shock-dominated OOD generalization performance of ICON ($\tau=0.1$) on the viscous Burgers and shallow water problems. Reported as mean \(\pm\) std over three training runs.}
\label{tab:icon_otherproblems}
\setlength{\tabcolsep}{8pt}
\renewcommand{\arraystretch}{1.15}
\resizebox{\textwidth}{!}{%
\begin{tabular}{lcccc}
\toprule
Dataset & \multicolumn{2}{c}{In-distribution} & \multicolumn{2}{c}{Shock IC} \\
\cmidrule(lr){2-3}\cmidrule(lr){4-5}
& Rel.~$l^2$ ($\downarrow$) & Rel.~$l^\infty$ ($\downarrow$) & Rel.~$l^2$ ($\downarrow$) & Rel.~$l^\infty$ ($\downarrow$) \\
\midrule
Shallow water & 3.15e-1$\pm$4.74e-3 & 9.16e-1$\pm$3.20e-3& 2.18e-1$\pm$3.23e-3  & 7.78e-1$\pm$4.97e-3  \\
\midrule
Viscous Burgers & 5.55e-2$\pm$7.48e-2  & 2.95e-1$\pm$2.86e-1 & 1.03e-2$\pm$1.1e-3  & 1.33e-1$\pm$7.25e-3  \\
\bottomrule
\end{tabular}%
}
\end{table*}

\section{Further architecture details}
\label{appendix: further architecture details}
Here, we provide additional details that supplement the main architecture description in \cref{section: model architecture}.

\subsection{Encoder internals and design rationale}
\label{appendix: encoder design}
The role of the context encoder is to compress a time-stacked series of spatial fields in to a 1D context vector. For time-varying solutions on a regular 2D grid, this problem strongly resembles that of video classification or regression tasks. Note that this is analogous to operator learning in 2D being similar to computer vision problems - indeed, many operator learning architectures have been inspired by vision models such as UNets or vision transformers (ViTs) based on this observation.

For our context encoder, we took inspiration from video vision transformer architectures. The classic model in this architecture class is the ViViT\footnote{With tubelet encoding and spatio-temporal attention to be precise. This is because the ViViT paper considers different embedding and attention strategies in their work.} \citep{arnab2021}, which extends the patch embedding strategy of the ViT to videos using spatio-temporal "tubelets". However, through ablation experiments (\cref{appendix: encoder ablation}), we found that basing the encoder on the more recent TRecViT model \citep{patraucean2025} gave better results.

\subsubsection{ViViT encoder (Ablation)}
ViViT \citep{arnab2021} naturally extends the ViT \citep{dosovitskiy2020} architecture by increasing the dimension of the patches to include a time dimension as well. Thus, for a \((s+1)\)-D spatiotemporal input \(\bm{u}^{n-k+1:n}\in \mathbb{R}^{k\times d\times N_{\bm{x}}}\), ViViT uses spatiotemporal patches (or `tubelets') \(P\in\Real^{N_{p_t}\times N_{\bm{p_x}}}\) (\(N_{\bm{x}}, N_{\bm{p_x}}\) denote \(N_{x_1}\times\cdots\times N_{x_s}, N_{{p_{x_1}}}\times\cdots\times N_{{p_{x_s}}}\), respectively), resulting in \(P:=\ceil{\frac{N_t}{N_{p_t}}}\times \ceil{\frac{N_{x_1}}{N_{p_{x_1}}}} \times \cdots \times \ceil{\frac{N_{x_s}}{N_{p_{x_s}}}}\) patches \footnote{The ceiling function appears because the input data is appropriately padded when the data dimensions are not strict multiples of the patch dimensions.}.

Subsequent calculations are identical to the standard ViT pipeline:
\begin{equation}
    \begin{aligned}
        &\bm{v}^{(0)}\in \Real^{P\times e} = \mathrm{PatchEmbed}(\bm{u}^{n-k+1:n})\\
        &\bm{v}^{(\ell+1)}\in \Real^{P\times e}=\mathrm{TransformerEncoder^{(\ell)}}(\bm{v}^{(\ell)}), \; \ell=0, \dots ,L-1\\
        &\bm{c}\in\Real^e = \frac{1}{P}\sum_{p=1}^{P}\mathrm{LayerNorm}(v^{(L)}_p).
    \end{aligned}
\end{equation}
Here, each input patch of shape (\(N_{p_t}, d, N_{\bm{p_x}})\) is first mapped to a \(e\)-dimensional latent vector, and the appropriate positional encoding vectors are added the result. The resulting patch embeddings \(\bm{v}^{(0)}\) are then passed though a depth-\(L\) transformer encoder, which repeatedly computes self-attention across the inputs to yield the output \(\bm{v}^{(L)}\). Finally, the context vector \(\bm{c}\) is obtained by applying LayerNorm across the channel dimension and then taking the mean across the patch dimension. Analogously to the design of DPOT \citep{hao2024}, we used a two layer MLP with GeLU nonlinearities to map the patches to latent vectors, and learned positional encodings as the position encoding strategy.

\subsubsection{TRecViT encoder}
For the in-context operator learning problem, the ideal encoder must respect causality of the input data. This is because the encoder must identify the relevant PDE parameters from the input context - that is, solve the inverse PDE problem - and the behavior of the solution over time provides key insights to the underlying system, such as wave propagation speed or dissipation.

While the ViViT encoder is an effective method to compress spatiotemporal data into a single context vector, it does not respect the causality of the input data. This is because ViViT treats the temporal and spatial axes identically, and scrambles the temporal and spatial information all at once via self-attention. 

Alternatively, TRecViT \citep{patraucean2025} is a recent ViT-based architecture, designed for causal video modeling. Causality of the input data is respected by treating the spatial and temporal dimensions separately: spatial axes are handled analogously to a standard ViT, while the temporal axis is processed by a state space model. Concretely, for a \((s+1)\)-D spatiotemporal input \(\bm{u}^{n-k+1:n}\in \mathbb{R}^{k\times d\times N_{\bm{x}}}\), TRecViT uses spatial patches \(\mathscr{P}\in\Real^{N_{\bm{p_x}}}=\Real^{N_{{p_{x_1}}}\times\cdots\times N_{{p_{x_s}}}}\) to produce \(P:=\ceil{\frac{N_{x_1}}{N_{p_{x_1}}}} \times \cdots \times \ceil{\frac{N_{x_s}}{N_{p_{x_s}}}}\) input patches per timestep. The resulting patches are then processed as,

\begin{equation}
    \begin{aligned}
        &\bm{v}^{(0)}\in \Real^{k\times n_P\times e} = \mathrm{PatchEmbed}(\bm{u}^{n-k+1:n})\\
        &\widehat{\bm{v}}^{(\ell)}_{:, p} \in \Real^{k\times e}=\mathrm{TemporalBlock^{(\ell)}}(\bm{v}^{(\ell)}_{:,p}), \; \ell=0, \dots ,L-1\\
        &\bm{v}^{(\ell+1)}_{t,:}\in \Real^{P\times e}=\mathrm{TransformerEncoder}^{(\ell)}(\widehat{\bm{v}}^{(\ell)}_{t,:}), \; \ell=0, \dots ,L-1\\
        &\bm{c}\in\Real^e = \frac{1}{P}\sum_{p=1}^{P}\mathrm{LayerNorm}(\bm{v}^{(L)}_{k,p}).
    \end{aligned}
\end{equation}

Here, we see that the spatially patched input context is processed in an alternating manner. First, for a given spatial patch index \(p\), the corresponding time evolution of the patches (or a `temporal tube') \(\bm{v}^{(\ell)}_{:,p}\) is processed by the TemporalBlock to yield a new patch sequence. This TemporalBlock is based on the gated recurrent unit \citep{de2024}, and mixes temporal information in a causal manner. Afterwards, the resulting modified patches are then mixed spatially for each time using a standard transformer encoder. To obtain a single representative context vector from the spatiotemporal output \(\bm{v}^{(L)}\), we first apply LayerNorm across the channel dimension, and average across the patch dimension at the final time step (\(t=k\)).


\subsection{Hypernetwork}

\paragraph{Design of the matrix \(W_{out}\)} The output layer weights \(W_{out}\) has the shape \((q,h)\), resulting in \(q\cdot h\) parameters when instantiated as a dense matrix. The problem here is that \(q\) is a very large number (\(\sim O(10^6)\) in our experiments), making the total number of parameters exceedingly large. To resolve this problem, we set \(W_{out}\) to be a block diagonal matrix with \(B\) blocks, \footnote{This is a widely used trick in the literature - e. g., \citet{guibas2021} for Fourier neural operators, and \citet{de2024} for state space models.} resulting in \((q\cdot h)/B\) parameters.

\paragraph{Parameter initialization of the hypernetwork} For hypernetworks, care needs to be taken when initializing its parameters. This is because unlike normal neural networks, where the goal of initialization is to place the neural network in a region of parameter space amenable to gradient descent, hypernetworks must be initialized so that its corresponding outputs are good initial parameters for the target network downstream. As such, applying conventional neural network initialization schemes to hypernetworks can result in target network initializations that are unstable to train \citep{chang2019,ortiz2023,beck2023}.

For the initialization of our hypernetwork, we chose the Bias-Hyperinit method \citep{beck2023} due to its ease of implementation and compatibility with arbitrary target network architectures. The Bias-Hyperinit method is based on the observation that if \(W_{out}=0\), then \(H(\bm{c})=\bm{b}_{out}\) - that is, the generated target network parameters become the output layer bias \(\bm{b}_{out}\), regardless of the hypernetwork input. Accordingly, the parameters of the hypernetwork are initialized as follows:
\begin{itemize}
    \item \(W_{in}, \bm{b}_{in}\): Initialized according to the default initialization strategy of the neural network library; e.g, Xavier initialization.
    \item \(W_{out}\): Initialized as a zero matrix.
    \item \(\bm{b}_{out}\): Set as a concatenated vector of the default initialization parameters of the target network.
\end{itemize}
Hypernetwork initialized this way outputs the initial parameters of the target network according to the target network's default initialization strategy, preventing training instabilities.

\section{Further experimental details}
\subsection{Model configuration}
\label{modelconfigs}
We chose architectural parameters for the different models so that all models would have comparable number of trainable parameters. For baseline models, this was done by starting with the configuration in the original code, then adjusting values to achieve the desired parameter count. Parameters that appear across multiple models, such as patch size (for patch embedding) or embedding dimension was set to be equal for fairer comparisons.

%

\begin{table*}[ht!]
\centering
\setlength{\tabcolsep}{5pt}
\renewcommand{\arraystretch}{1.15}

\begin{minipage}[t]{0.48\textwidth}
\centering
\caption{Configuration for DPOT}
\begin{tabularx}{\linewidth}{YC}
\toprule
Name & Value (1D / 2D) \\
\midrule
Patch size & [4] / [8,8] \\
Lift dim & 128 \\
Layers & 10 \\
Heads & 8 \\
MLP hidden dim & 512 \\
Frequency modes & [12] / [10,10] \\
\bottomrule
\end{tabularx}
\label{tab:dpot_config}
\end{minipage}
\hfill
\begin{minipage}[t]{0.48\textwidth}
\centering
\caption{Configuration for ICON}
\begin{tabularx}{\linewidth}{YC}
\toprule
Name & Value \\
\midrule
Condition length & 100 \\
QoI length & 100 \\
Number of demos & 20 \\
Hidden dim & 256 \\
Layers & 6 \\
Heads & 8 \\
\bottomrule
\end{tabularx}
\label{tab:icon_config}
\end{minipage}

\vspace{1.5em}

\begin{minipage}[t]{0.48\textwidth}
\centering
\caption{Configuration for DISCO}
\begin{tabularx}{\linewidth}{cYC}
\toprule
& Name & Value (1D / 2D) \\
\midrule
\multirow{5}{*}{\rotatebox[origin=c]{90}{Encoder}}
& Patch size & [4] / [8,8] \\
& Embedding dim & 128 \\
& Time-space blocks & 3 \\
& Heads & 8 \\
& MLP hidden dim & 32 \\
\midrule
\multirow{5}{*}{\rotatebox[origin=c]{90}{Target}}
& Channels & 4 / 8 \\
& GroupNorm groups & 2 / 4 \\
& ODE solve rel. tolerance & $10^{-4}$ \\
& ODE solve abs. tolerance & $10^{-6}$ \\
& Max. solver steps & 16 \\
\bottomrule
\end{tabularx}
\label{tab:disco_config}
\end{minipage}
\hfill
\begin{minipage}[t]{0.48\textwidth}
\centering
\caption{Configuration for HFluxNO}
\begin{tabularx}{\linewidth}{cYC}
\toprule
& Name & Value (1D / 2D) \\
\midrule
\multirow{7}{*}{\rotatebox[origin=c]{90}{Encoder}}
& Patch size & [4] / [8,8] \\
& Embedding dim & 128 \\
& Layers & 2 \\
& Recurrent unit width & 128 \\
& Heads & 8 \\
& MLP hidden dim & 64 \\
& Blocks (\(B\)) & 8 \\
\midrule
\multirow{4}{*}{\rotatebox[origin=c]{90}{Target}}
& Lift dim & 128 \\
& Stencil size & [10,10] / [8,8] \\
& Layers & 4 / 5 \\
& MLP hidden dim & 128 \\
\bottomrule
\end{tabularx}
\label{tab:hfluxno_config}
\end{minipage}

\end{table*}
\section{Data generation}
\label{appendix: data generation}

We generate numerical trajectories using classical finite-volume or finite-difference solvers and use them as supervised training data. For each equation family, we sample equation parameters from a prescribed distribution and independently sample initial conditions. Each pair of equation parameters and initial condition defines one trajectory. The equation parameters are stored as metadata but are not provided as model inputs during training or evaluation.

Unless otherwise stated, all simulations are performed on the periodic spatial domain \(x\in[0,1]\). For all newly generated 1D datasets, we use \(N_x=100\) spatial grid cells and save \(N_t=100\) time snapshots. We sample \(1000\), \(100\), and \(100\) coefficient choices for the training, validation, and test datasets, respectively. For each coefficient choice, we generate \(100\) independent initial conditions. This results in \(100{,}000\) training trajectories, \(10{,}000\) validation trajectories, and \(10{,}000\) test trajectories.

\subsection{1D Cubic Conservation Laws}
\label{appendix:cubic-data-generation}

We consider the one-dimensional scalar conservation law
\[
    u_t + f(u)_x = 0,
    \qquad x\in[0,1],
\]
with periodic boundary conditions. The flux is given by
\[
    f(u)=a u^3+b u^2+c u,
    \qquad
    (a,b,c)\sim \mathrm{Unif}([-1,1]^3).
\]
Initial conditions are sampled from a mean-zero periodic Gaussian random field with covariance kernel
\[
    k(x,x')
    =
    \exp\left(
        -\left(1-\cos(2\pi(x-x'))\right)
    \right).
\]

The trajectories are generated using \texttt{PyClaw} \citep{pyclaw} with a custom scalar Riemann solver for the cubic flux. We use the MC total-variation-diminishing limiter, one wave family, desired CFL number \(0.5\), and maximum CFL number \(0.9\). The resulting dataset has shape
\[
    [N_c,N_{\mathrm{init}},N_t,N_x,N_q]
    =
    [1000,100,100,100,1]
\]
for the training split, where \(N_q=1\) is the number of conserved variables.

\paragraph{OOD dataset simulations} For the out-of-distribution experiments, we consider both different types of initial conditions and different equation forms.

The shock dominated initial conditions were generated by generating random periodic step functions with variable number of steps and step heights. The minimum and maximum number of steps were set to 1 and 5 respectively, and the minimum and maximum step heights were set to -1 and 1.

The different equation form considered was the sine flux-based conservation law, with the flux given as,
\[
    f(u)=a\sin(bu),
    \qquad
    (a,b)\sim \mathrm{Unif}([-1,1]^3).
\]
For all three types of OOD datasets created (different initial conditions, different equations, different initial conditions and equations), we also sampled 100 coefficient choices and 100 initial conditions per coefficient choice, resulting in 10,000 test trajectories.

\subsection{1D Parametric Shallow Water Equations}
\label{appendix:shallow-water-data-generation}

We consider a two-component parametric shallow-water-type conservation law. The state is
\[
    q=(h,m)^\top,
    \qquad m=hu,
\]
where \(h\) is the water height and \(m\) is the momentum. The governing equation is
\[
    q_t + F(q)_x = 0,
    \qquad x\in[0,1],
\]
with flux
\[
    F(q)
    =
    \begin{pmatrix}
        \alpha m \\
        \gamma m^2/h + \frac{1}{2}\beta h^2
    \end{pmatrix}.
\]
The parameters are sampled as
\[
    (\alpha,\gamma,\beta)
    \sim
    \mathrm{Unif}
    \left(
        [0.5,1.5]\times[0.5,1.5]\times[8,12]
    \right).
\]
The standard shallow water equations correspond to \((\alpha,\gamma,\beta)=(1,1,g)\).

For the initial conditions, we sample \(m(0)\) from a Gaussian random field with a Gaussian covariance function
\[
    k(x,x')
    =
    \sigma^2\left(1-\exp\left(
        -\frac{s^2|x-x'|^2}{l^2}
    \right)\right),
\]
where we used \(\sigma^2=0.5\) and \(l=0.3\). 

To ensure positivity of the water height, \(h(0)\) is sampled from a lognormal random field, whose covariance function was set identical to that of \(m(0)\). In the numerical solver, a small height floor \(h_{\mathrm{floor}}=10^{-8}\) is used for divisions and square roots. The Gaussian and lognormal random fields on a periodic lattice was generated using the \texttt{gstools} \citep{muller2022} package.

Trajectories are generated using \texttt{PyClaw} \citep{pyclaw} with a custom Roe-type approximate Riemann solver. We use the MC total-variation-diminishing limiter, two wave families, desired CFL number \(0.5\), and maximum CFL number \(0.9\). The resulting dataset has two state channels, corresponding to \(h\) and \(m\), and hence has shape
\[
    [N_c,N_{\mathrm{init}},N_t,N_x,N_q]
    =
    [1000,100,100,100,2]
\]
for the training split.

\paragraph{OOD dataset simulations} For the out-of-distribution experiment, we generated a dataset with different initial condition family for \(h(0)\): shock-dominated versions of \(h(0)\) were generated using random periodic step functions as for the cubic conservation law case. The minimum and maximum number of steps were set to 1 and 5 respectively, and the minimum and maximum step heights were set to 0.5 and 4.5 to abide by the non-negativity constraint. The initial conditions for \(m(0)\) was kept identical to the in-distribution case, as we found that using step functions for both field results in excessively irregular solutions. Likewise, we sampled 100 coefficient choices and 100 initial conditions per coefficient choice, resulting in 10,000 test trajectories.

\subsection{1D Viscous Burgers Equation}
\label{appendix:viscous-burgers-data-generation}

We consider the parametric viscous Burgers-type equation
\[
    u_t + a(u^2)_x = b u_{xx},
    \qquad x\in[0,1],
\]
with periodic boundary conditions. The parameters are sampled as
\[
    (a,b)
    \sim
    \mathrm{Unif}
    \left(
        [0.5,1.5]\times[0.005,0.015]
    \right).
\]
This dataset contains an explicit diffusion term and is included to test whether the proposed architecture can handle dynamics beyond strictly hyperbolic conservation laws.

Initial conditions are sampled from the same class of one-dimensional Gaussian random fields used for the scalar conservation-law experiments. The equation is solved using an explicit finite-volume/finite-difference scheme: the nonlinear advective term is discretized with a local Rusanov flux, while the diffusion term is discretized using a centered second-order finite difference. Periodic boundary conditions are imposed throughout the simulation.

The internal time step is chosen adaptively using both advective and diffusive stability constraints, with CFL number \(0.4\). The solver is forced to land exactly on each saved output time by shortening the final internal step before a saved snapshot if necessary. The resulting dataset has shape
\[
    [N_c,N_{\mathrm{init}},N_t,N_x,N_q]
    =
    [1000,100,100,100,1]
\]
for the training split.

\paragraph{OOD dataset simulations} For the out-of-distribution experiment, we generated a shock-dominated initial condition dataset. The initial conditions were once again set to periodic random step functions, whose parameters were identical to the cubic conservation law case. Once again, we sampled 100 coefficient choices and 100 initial conditions per coefficient choice, resulting in 10,000 test trajectories.

\section{Ablation study}
\subsection{Encoder ablation}
\label{appendix: encoder ablation}

To validate our architectural decision of basing the encoder on the TRecViT model \citep{patraucean2025} instead of the more simpler ViViT \citep{arnab2021} (see \cref{appendix: encoder design} for details on the encoder architectures), we trained a version of our HFluxNO model using ViViT as the encoder architecture on the 1D cubic conservation law dataset. The encoder parameters used are shown in \cref{tab:trecvit encoder config,tab:vit encoder config}. These were chosen so that the different encoders have similar number of trainable parameters and common parameters have identical values.

\begin{table}[h!]
\setlength{\tabcolsep}{7pt}
\renewcommand{\arraystretch}{1.15}

\begin{minipage}{0.5\linewidth}
\centering
\caption{Recurrent ViT Encoder}
\begin{tabular}{lc}
\toprule
Name & Value \\
\midrule
Patch size (\(x\)) & [4] \\
Embedding dim & 128 \\
Layers & 2 \\
Recurrent unit width & 128 \\
Heads & 8 \\
MLP hidden dim & 64 \\
Blocks & 8 \\
\midrule
Encoder parameters & 279424 \\
Total parameters & 2678020 \\
\bottomrule
\end{tabular}
\label{tab:trecvit encoder config}
\end{minipage}
\begin{minipage}{0.5\linewidth}
\centering
\caption{ViViT Encoder}
\begin{tabular}{lc}
\toprule
Name & Value \\
\midrule
Patch size (\(t, x\)) & [4, 4] \\
Embedding dim & 128 \\
Layers & 3 \\
Heads & 8 \\
MLP hidden dim & 64 \\
\midrule
Encoder parameters & 268864 \\
Total parameters & 2667460 \\
\bottomrule
\end{tabular}
\label{tab:vit encoder config}
\end{minipage}
\end{table}

From the results shown in \cref{appendix: encoder ablation}, we find that the TRecViT encoder is indeed stronger than the ViViT encoder. Comparing against the baseline results (\cref{tab: cubid_1d_in_distribution}), we find that the performance from the ViViT encoder is similar to the runner up models (DPOT for Rel.~\(l^2\), DISCO for Rel.~\(l^\infty\)). 

\begin{table*}[htp!]
\centering
\caption{Encoder ablation results for the 1D cubic conservation law dataset.
Reported as mean \(\pm\) std over three training runs. The best results are in bold.}
\label{tab: encoder ablation}
\setlength{\tabcolsep}{7pt}
\renewcommand{\arraystretch}{1.15}
\resizebox{\textwidth}{!}{\begin{tabular}{llcccc}
\toprule
& & \multicolumn{2}{c}{Single step} & \multicolumn{2}{c}{Autoregressive Rollout (20 steps)} \\
\cmidrule(lr){3-4}\cmidrule(lr){5-6}
Dataset & Encoder & Rel.~$l^2\ (\downarrow)$  & Rel.~$l^\infty\ (\downarrow)$ & Rel.~$l^2\ (\downarrow)$  & Rel.~$l^\infty\ (\downarrow)$\\
\midrule
\multirow{3}{*}{Cubic} & ViViT & 7.14e-3$\pm$8.07e-4 & 2.57e-2$\pm$2.86e-3 & 7.18e-2$\pm$5.21e-3 & 4.28e-1\(\pm\)1.63e-2\\

\cmidrule(lr){2-6}
& TRecViT & \textbf{4.10e-3$\pm$1.15e-4} & \textbf{1.55e-2$\pm$6.17e-4} & \textbf{5.21e-2\(\pm\)2.15e-3} & \textbf{3.68e-1\(\pm\)1.56e-2} \\

\bottomrule
\end{tabular}}
\end{table*}

\subsection{Target network ablation}

\begin{table}[h!]
\setlength{\tabcolsep}{2pt}
\renewcommand{\arraystretch}{1.15}

\begin{minipage}{0.33\linewidth}
\centering
\caption{FNO Target}
\begin{tabular}{lc}
\toprule
Name & Value \\
\midrule
Lift dim & 48 \\
Frequency modes & 8\\
Layers & 4 \\
MLP hidden dim & 48 \\
\midrule
Target params (\(q\)) & 88033 \\
Total params & 3217636 \\
\bottomrule
\end{tabular}
\label{tab:fno target network config}
\end{minipage}
\begin{minipage}{0.33\linewidth}
\centering
\caption{UNet Target}
\begin{tabular}{lc}
\toprule
Name & Value \\
\midrule
Channels & 8\\
GroupNorm groups & 4 \\
\midrule
Target params (\(q\))  & 112081 \\
Total params & 4011220 \\
\bottomrule
\end{tabular}
\label{tab:unet target network config}
\end{minipage}
\begin{minipage}{0.33\linewidth}
\centering
\caption{Flux NO Target}
\begin{tabular}{lc}
\toprule
Name & Value \\
\midrule
Lift dim & 128 \\
Stencil size & [10, 10] \\
Layers & 4 \\
MLP hidden dim & 128 \\
\midrule
Target params (\(q\))  & 71681 \\
Total params & 2678020 \\
\bottomrule
\end{tabular}
\label{tab:fluxno target network config}
\end{minipage}
\end{table}

\begin{table*}[htp!]
\centering
\caption{Target network ablation results for the 1D cubic conservation law dataset.
Reported as mean \(\pm\) std over three training runs. The best results are in bold.}
\label{tab: target network ablation}
\setlength{\tabcolsep}{7pt}
\renewcommand{\arraystretch}{1.15}
\resizebox{\textwidth}{!}{\begin{tabular}{llcccc}
\toprule
& & \multicolumn{2}{c}{Single step} & \multicolumn{2}{c}{Autoregressive Rollout (20 steps)} \\
\cmidrule(lr){3-4}\cmidrule(lr){5-6}
Dataset & Target & Rel.~$l^2\ (\downarrow)$  & Rel.~$l^\infty\ (\downarrow)$ & Rel.~$l^2\ (\downarrow)$  & Rel.~$l^\infty\ (\downarrow)$\\
\midrule
\multirow{3}{*}{Cubic} & FNO & 1.61e-2$\pm$5.24e-4 & 5.65e-2$\pm$2.30e-3 & 1.53e-1$\pm$5.03e-3 & 7.21e-1\(\pm\)1.22e-2\\
& UNet & 7.49e-3$\pm$5.54e-4 & 2.74e-2$\pm$2.37e-3 & 8.30e-2\(\pm\)7.01e-3 & 5.13e-1\(\pm\)3.49e-2\\
\cmidrule(lr){2-6}
& Flux NO & \textbf{4.10e-3$\pm$1.15e-4} & \textbf{1.55e-2$\pm$6.17e-4} & \textbf{5.21e-2\(\pm\)2.15e-3} & \textbf{3.68e-1\(\pm\)1.56e-2} \\

\bottomrule
\end{tabular}}
\end{table*}

\end{document}